\documentclass[10pt,twocolumn,letterpaper]{article}

\usepackage{iccv}
\usepackage{times}
\usepackage{epsfig}
\usepackage{graphicx}
\usepackage{amsmath}
\usepackage{amssymb}
\usepackage{subfigure}
\usepackage{float}
\usepackage{multirow}

\graphicspath{{pics/}}

\usepackage[pagebackref=true,breaklinks=true,letterpaper=true,colorlinks,bookmarks=false]{hyperref}

\iccvfinalcopy 

\def\iccvPaperID{xxxx} 

\begin{document}

\title{CenterAtt: Fast 2-stage Center Attention Network \\ A solution for Waymo Open Dataset Real-time 3D Detection Challenge}


\author{
Jianyun Xu
\quad
Xin Tang
\quad
Jian Dou
\quad
Xu Shu
\quad
Yushi Zhu
\\
Hikvision Research Institute
\\
{\tt\small \{xujianyun, tangxin10, doujian, shuxu, zhuyushi\}@hikvision.com}
}



\maketitle
\ificcvfinal\thispagestyle{empty}\fi

\begin{abstract}
   In this technical report, we introduce the methods of \textbf{HIKVISION\_LiDAR\_Det} in the challenge of waymo open dataset real-time 3D detection. Our solution for the competition are built upon Centerpoint 3D detection framework. Several variants of CenterPoint are explored, including center attention head and feature pyramid network neck. In order to achieve real time detection, methods like batchnorm merge, half-precision floating point network and GPU-accelerated voxelization process are adopted. By using these methods, our team ranks 6th among all the methods on real-time 3D detection challenge in the waymo open dataset.
\end{abstract}


\section{Introduction}
The Waymo Open Dataset Challenges at CVPR’21 are the highly competitive competition with the largest LiDAR point cloud and camera dataset for autonomous driving. We mainly focus on the realtime 3D detection task, which requires the algorithm to detection objects as a set of 3D bounding boxes, within a time consumtipn limited up to 70 ms.

\begin{figure*}
    \centering
    \includegraphics[width=0.93\linewidth]{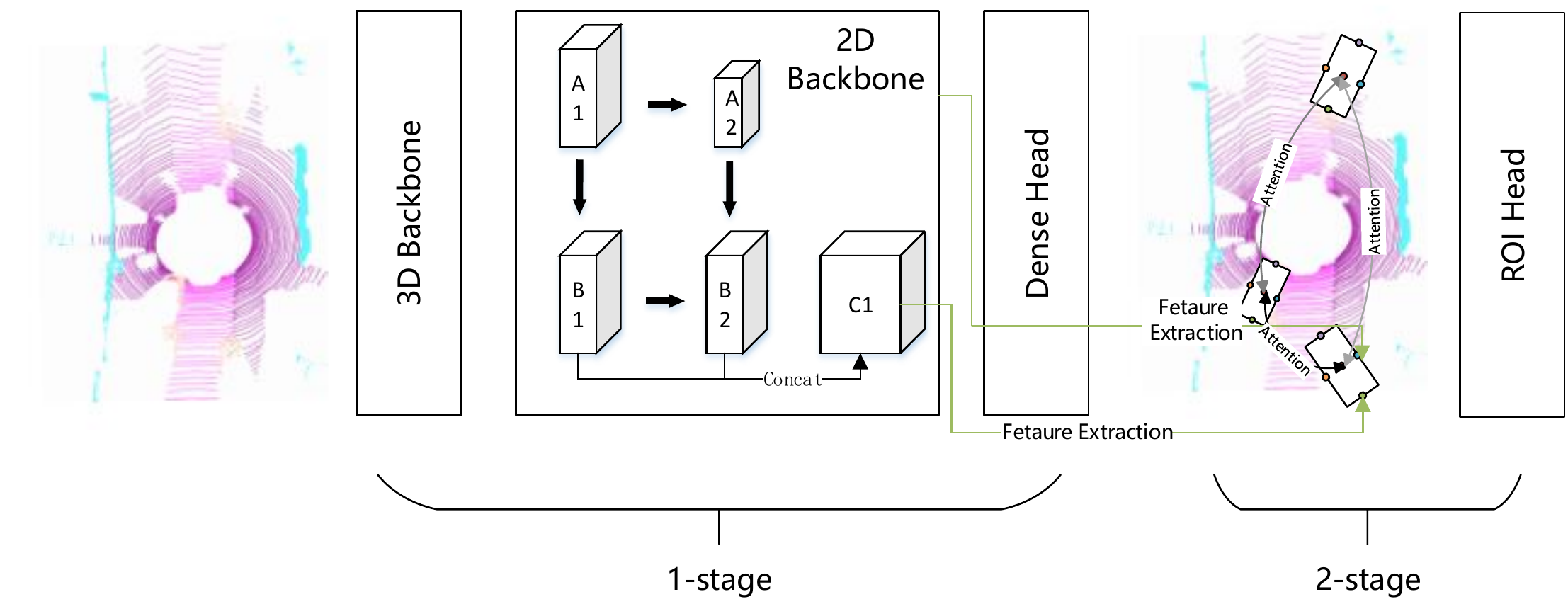}
    \caption{Overview framework.}
    \label{fig:framework}
\end{figure*}

\section{Methodology}
\label{sec:Methodology}
We explore our methods based on CenterPoint~\cite{yin2020center}. In this section, we first review the CenterPoint and then introduce two improvements on it.

\subsection{CenterPoint Review}
Two-Stage CenterPoint~\cite{yin2020center} detects centers of objects using a keypoint detector and regresses to other attributes, including 3D size, 3D orientation. The first stage of CenterPoint predicts a class-specific dense heatmap, object size, a sub-voxel location refinement and rotation. The center-head’s goal is to produce a heatmap peak at the center location of any detected object. This head produces a K-channel heatmap, and K denotes the number of classes. During training, it targets a 2D Gaussian produced by the projection of 3D centers of annotated bounding boxes into the bird eye view. At inference time, CenterPoint extracted all properties by indexing into dense regression head outputs at each object’s peak location. In the second stage, it refines these estimates by using additional point features on the object. CenterPoint extracted one point-feature from the 3D center of each face of the predicted bounding box and only considered the four outward-facing box-faces together with the predicted object center. For each point, The features are extracted by using bilinear interpolation from the intermediate backbone’s feature map. Then the extracted point-features are concatenate and passed through an MLP. The second stage predicts a class-agnostic confidence score and box refinement on top of one-stage CenterPoint’s prediction results. The final confidence score is computed as the geometric average of the two score which includes the class prediction from one-stage CenterPoint and a class-agnostic confidence score from the second-stage. For box regression, the model predicts a refinement on top of first stage proposals.

\subsection{Center Attention Head} 
Original CenterPoint~\cite{yin2020center} does not carry out any post-processing in the second stage, and keeps many low-score and overlapped bounding boxes in the final output, which is not appropriate in the realistic applications. We tried some score filter or NMS thresholds to handle this problem, but most of them would decrease the mAP performance, which indicates that there exists many boxes that have lower scores but higher IoUs with the ground truth. This inspired us to solve this problem more elegantly, and we struggled to employ the self-attention network in the second stage.

Overview framework is shown in Figure~\ref{fig:framework}, this section is mainly about 2-stage. Firstly, we use 1-stage output boxes as proposals and apply the ROI pooling same as CenterPoint~\cite{yin2020center} which is competitively efficient. Then these ROI features along with their position encodings are feed into a multi-head self-attention layer, and the proposals would interact with each other. Finally, two light head, i.e. classification and regression head, are connected to get the final output. In detail, we use an 8-head self-attention layer with 2048 feed forward dimension, and apply sine position embedding with 128 output channels.

As for the loss, we use Hungarian algorithm to find a bipartite matching between ground-truth and proposals, and the matching cost is defined as follows:
\begin{equation}\label{eq: cost matrix}
    \mathcal{L}=\lambda_{cls}\mathcal{L}_{cls}+\lambda_{iou}\mathcal{L}_{iou}
\end{equation}
where $\mathcal{L}_{cls}$ is the classification cost to prevent different classes match together, and $\mathcal{L}_{iou}$ is rotated IoU between proposals and ground-truth. $\lambda_{cls}$ and $\lambda_{iou}$ are coefficients of each components. After matching, the normal binary cross entropy and L1 loss are applied for classification and regression, respectively.

\subsection{FPN Neck} 
In the second stage of original CenterPoint~\cite{yin2020center}, it extracts additional ROI features from the single scale feature map, which may result in loss of information for small objects like cyclist and pedestrian. Inspired by FPN~\cite{FPN}, we use multi scale feature maps for ROI feature extraction, as depicted in Figure~\ref{fig:framework} 2D Backbone. Specifically, we project object center and center of each face of the predicted bounding box to different feature maps with respective stride and extract a feature using bilinear interpolation. All point features of a proposal are concatenated to form the proposal-wise feature and passed through an MLP.

\section{Latency Analysis}
\label{sec:Latency}
Latency also plays an essential role in this challenge. We first analyse the time cost of the original CenterPoint~\cite{yin2020center}, then tried to optimize some time-consuming modules, finally choose an appropriate baseline for further study.

We split the overall inference procedure into 5 steps: data load, preprocess, collate, load to GPU and model. Unsurprisingly, the original CenterPoint can not reach the latency requirement, and we tried to cut the backbone, put the voxelization to GPU, merge the BN to weight parameters, and use half-precision for inference. This process is showed in Table~\ref{tab:latency}.

We took the forth row(\textit{+ put voxelization to GPU}) as our baseline, and explore the above two methods (Sec.~\ref{sec:Methodology}) on it.


\begin{table*}[]
\begin{center}
\resizebox{0.9\textwidth}{!}{%
\begin{tabular}{l|rrrrrr|r}
\hline
\multicolumn{1}{c|}{\multirow{2}{*}{methods}} &
  \multicolumn{6}{c|}{latency(ms)} &
  \multicolumn{1}{c}{\multirow{2}{*}{mAPH}} \\
\multicolumn{1}{c|}{} &
  \multicolumn{1}{c}{load data} &
  \multicolumn{1}{c}{preprocess} &
  \multicolumn{1}{c}{collate} &
  \multicolumn{1}{c}{load to GPU} &
  \multicolumn{1}{c}{model} &
  \multicolumn{1}{c|}{overall} &
  \multicolumn{1}{c}{} \\ \hline
CenterPoint-1stage        & 10.0 & 17.5 & 1.0 & 22.7 & 65.8 & 117.0 & 66.3 \\
+ 2stage                  & 10.0 & 17.5 & 1.0 & 22.7 & 70.0 & 121.2 & 68.3 \\
+ backbone cut            & 10.0 & 17.5 & 1.0 & 22.7 & 51.0 & 102.2 & 67.5 \\
+ put voxelization to GPU & 10.0 & 0.0  & 1.0 & 1.0  & 52.9 & 64.9  & 67.5 \\ \hline
+ merge BN to weights     & 10.0 & 0.0  & 1.0 & 1.0  & 51.4 & 63.4  & 67.5 \\
+ half-precision          & 10.0 & 0.0  & 1.0 & 1.0  & 40.7 & 52.7  & 66.4 \\ \hline
\end{tabular}%
}
\end{center}
\caption{Latency Analysis. The mAPH is evaluated on 1/20 validation set. The performance drop happens when cut the backbone and use half-precision. We took the forth row(+ put voxelization to GPU) as our baseline.}
\label{tab:latency}
\end{table*}

Note that, we also tried to 
\begin{itemize}
    \item  decrease voxel resolution, but it turned out huge performance drop, so we choose to optimize the voxelization and keep original voxel resolution. 
    \item  use multi-frames, and put the point transformation into GPU. Despite we can run less than 70ms in our V100 server, we got about 77ms on the official server.
    \item  only use top lidar, but we cannot observe any latency earnings on the officail server, maybe due to the fluctuation of load.
\end{itemize}

\section{Experiments}
\subsection{Dataset}
For real-time 3D detection, Waymo Open Dataset~\cite{sun2020scalability} contains 798 training sequences and 202 validation sequences and 150 testing sequences. In terms of lidar data, it has five types of lidar, including top lidar, front lidar, side left lidar, side right lidar and rear lidar, and each of them has two returns. By default, we use all five lidars for training and testing. In order to meet the latency requirement, we just use a single frame of point cloud.

\subsection{Training and Inference}
We implement our detection model based on centerpoint~\cite{yin2020center}. We train our detector within range [-75.2m, 75.2m]for the X and Y axis, and [-2m, 4m] for the Z axis, and test in all ranges. Our model uses a [0.1m, 0.1m, 0.15m] voxel size, and keep all points if they fall into the same voxel and then average them as the voxel feature. 
Following centerpoint~\cite{yin2020center}, we use the same data augmentation strategy to improve the performance of our model. We flip lidar data along the X-axis and Y-axis randomly, and use random factors in [0:95, 1:05] for global scaling. We also rotate input lidar data with a random angle from [${-\pi/4}$, ${\pi/4}$]. In order to make the data distribution more balanced, we also use ground truth sampling, which copy and paste points from one frame to another from an annotated box.

We trained our model with batch size 32 on 8 V100s. The optimizer of our model is  adamw with one cycle strategy, setting the maximum learning rate to 1e-3, division factor to 10. Momentum range is 0.95-0.85, and weight decay is 0.01.

In order to verify our idea quickly and reduce the training time, we firstly use full data to train a 36 epochs one stage single frame centerpoint model as our pre-trained model, and test on 1/20 validation data. Then we finetune our models on this pre-trained model using 3 epochs or 6 epochs. In the final submission, since our model is a lightweight two-stage centerpoint variant, we will not freeze the weight when we use this finetune model. To improve the reasoning speed of our model, we also merge BatchNorm parameters to weight parameters, and use half-precision.

\subsection{Results}
\begin{figure*}[t]
\centering

\subfigure[original CenterPoint]{
    \begin{minipage}{0.47\linewidth}
    \includegraphics[width=1.1\linewidth]{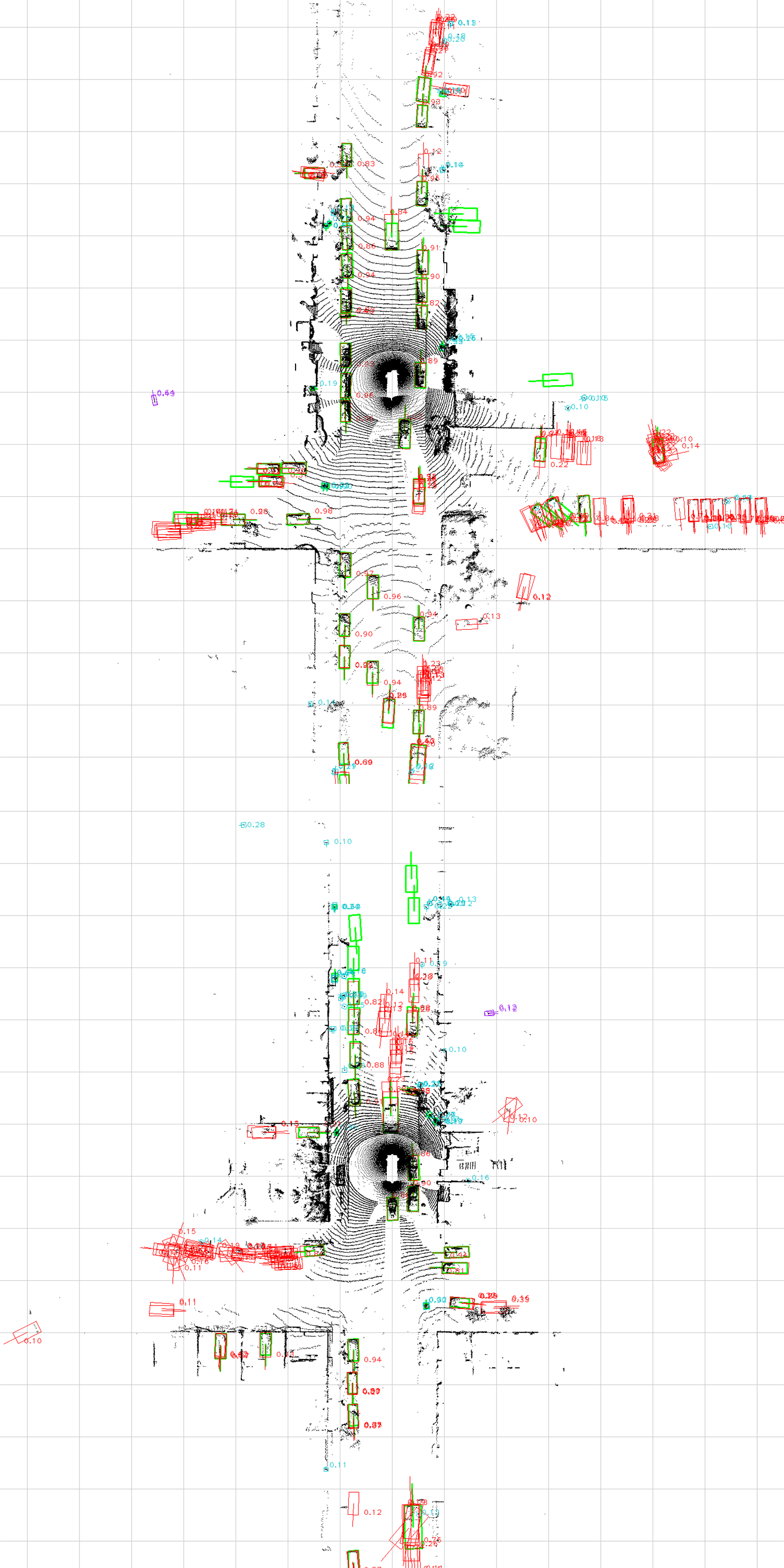}
    \end{minipage}
}
\subfigure[CenterAtt]{
    \begin{minipage}{0.47\linewidth}
    \includegraphics[width=1.1\linewidth]{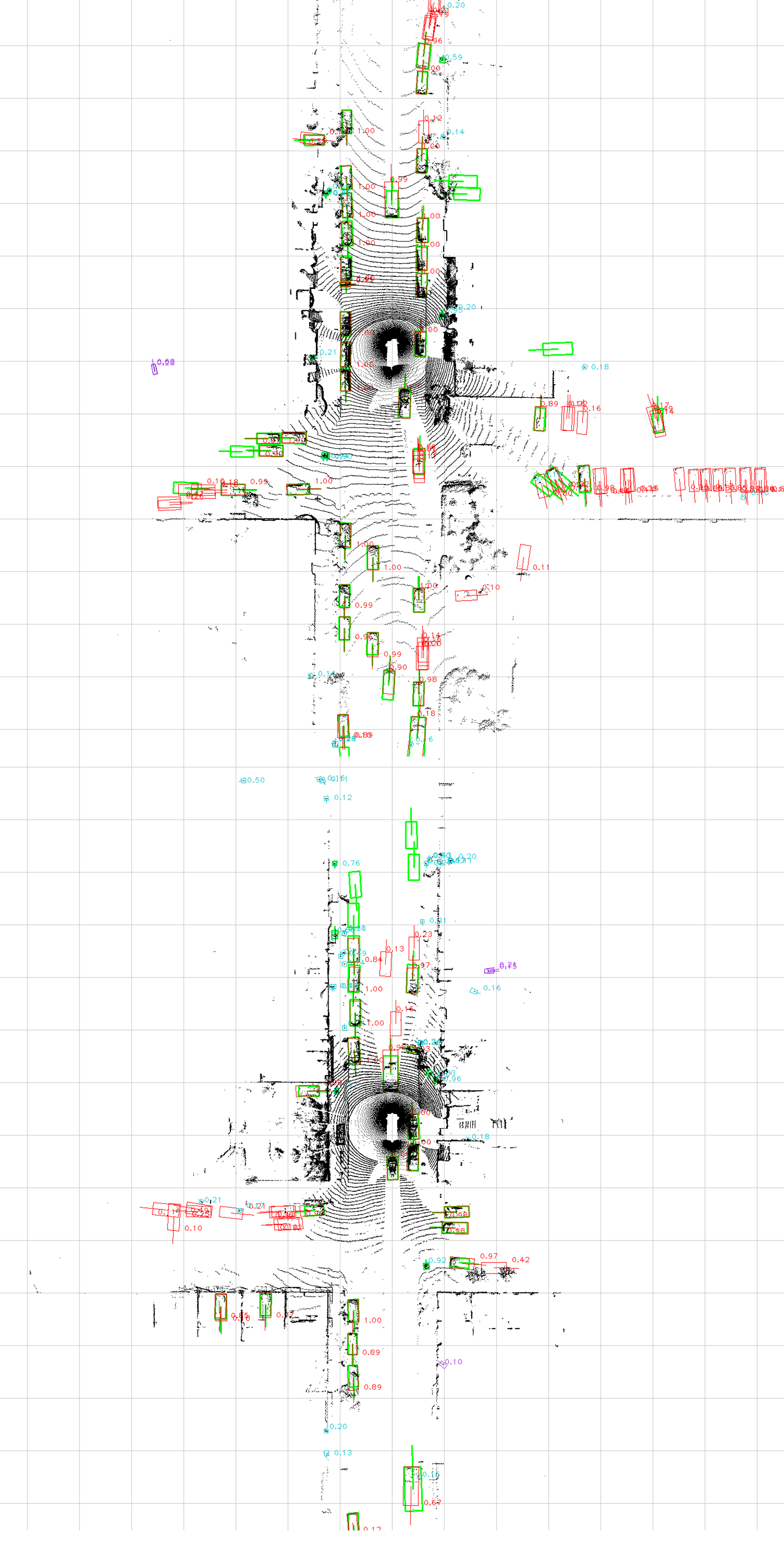}
    \end{minipage}
}

\caption{visualization: CenterPoint vs CenterAtt. Both side are filtered by 0.1 score. Best viewed in color. Green box indicates groud-truth, red indicates Vehicle, blue Pedestrian and purple Cyclist.}
\label{fig:vis}
\end{figure*}

Based on the baseline mentioned in Sec.~\ref{sec:Latency}, we tried our two methods proposed in Sec.~\ref{sec:Methodology}, and results are listed in Table~\ref{tab:component}. From this table, we can see very slight improvement over the baseline, but by employing CenterAtt head, the final output is more capable for realistic application, as shown in Figure~\ref{fig:vis}.

And the final submitted result is CenterAtt without FPN, by adding BN-merging and half-precision, the final results are listed in Table~\ref{tab:final}.

\begin{table}[]
\begin{center}
\resizebox{1.0\linewidth}{!}{%
\begin{tabular}{ccc|cc}
\hline
Baseline & CenterAtt Head & FPN Neck & latency(ms) & mAPH(\%) \\ \hline
$\surd$  &                &          &   64.9      &  67.5   \\
$\surd$  &  $\surd$       &          &   65.4      &  67.7   \\
$\surd$  &                & $\surd$  &   66.0      &  67.6   \\
$\surd$  &  $\surd$       & $\surd$  &   66.5      &  67.7   \\
\hline
\end{tabular}%
}
\end{center}
\caption{Ablation results. The mAPH is evaluated on 1/20 validation set.}
\label{tab:component}
\end{table}


\begin{table}[]
\begin{center}
\resizebox{1.0\linewidth}{!}{%
\begin{tabular}{l|lll|ll}
\hline
split & Veh\_L2 & Ped\_L2 & Cyc\_L2 & mAPH(\%) & Latency(ms) \\ \hline
val   & 66.9    & 62.7    & 67.6    & 65.7     & 55.3        \\
test  & 70.6    & 64.1    & 67.2    & 67.3     & 54.1    \\ \hline   
\end{tabular}%
}
\end{center}
\caption{CenterAtt final result on validation and test set.}
\label{tab:final}
\end{table}


\section{Acknowledgements}
Much thanks to the open-source code: OpenPCDet\footnote{https://github.com/open-mmlab/OpenPCDet}, CenterPoint\footnote{{https://github.com/tianweiy/CenterPoint?utm\_source=catalyzex.com}}, TorchSparse\footnote{https://github.com/mit-han-lab/torchsparse}.


{\small
\bibliographystyle{ieee_fullname}
\bibliography{egbib}
}

\end{document}